\documentclass[runningheads]{llncs}

\usepackage{eccv}

\usepackage{eccvabbrv}

\usepackage{graphicx}
\usepackage{booktabs}
\usepackage{adjustbox}
\usepackage{tabularx}
\usepackage{wrapfig}
\usepackage{multirow}

\usepackage[accsupp]{axessibility}  

\usepackage[pagebackref,breaklinks,colorlinks,citecolor=eccvblue]{hyperref}

\usepackage{orcidlink}

\begin{document}

\title{A Benchmark and Multi-Agent System for Instruction-driven Cinematic Video Compilation} 

\titlerunning{Instruction-driven Cinematic Video Compilation}


\author{
Peixuan Zhang\inst{\#1} \and Chang Zhou\inst{\dagger2} \and Ziyuan Zhang\inst{3} \and Hualuo Liu\inst{2} \and \\ Chunjie Zhang\inst{2} \and Jingqi Liu\inst{4,7} \and Xiaohui Zhou\inst{2} \and Xi Chen\inst{2} \and \\ Shuchen Weng\inst{*4,5} \and Si Li\inst{*1} \and Boxin Shi\inst{5,6}
}


\authorrunning{Zhang et al.}




\makeatletter
\newcommand{\samelineand}{\qquad \stepcounter{@inst}\textsuperscript{\the@inst}\enskip}
\makeatother

\institute{
\scriptsize 
School of Artificial Intelligence, Beijing University of Posts and Telecommunications \and
AI Technology Center, Online Video Business Unit, Tencent PCG \and
Tsinghua University \samelineand Beijing Academy of Artificial Intelligence \and
State Key Lab of Multimedia Info. Processing, School of Computer Science, Peking University \and
Nat’l Eng. Research Ctr. of Visual Technology, School of Computer Science, Peking University \and
School of Software and Microelectronics, Peking University \\[3pt]
\small
\email{\{pxzhang, lisi\}@bupt.edu.cn, chanzhou@tencent.com} 
}


\maketitle


{\let\thefootnote\relax\footnotetext{
$^{\#}$ Work done during an internship at Tencent PCG. \\
$^{\dagger}$ Project leader. \\
$^{*}$ Corresponding authors. \\

}}

\begin{abstract}
The surging demand for adapting long-form cinematic content into short videos has motivated the need for versatile automatic video compilation systems. 
However, existing compilation methods are limited to predefined tasks, and the community lacks a comprehensive benchmark to evaluate the cinematic compilation. 
To address this, we introduce CineBench, the first benchmark for instruction-driven cinematic video compilation, featuring diverse user instructions and high-quality ground-truth compilations annotated by professional editors. 
To overcome contextual collapse and temporal fragmentation, we present CineAgents, a multi-agent system that reformulates cinematic video compilation into ``design-and-compose'' paradigm. CineAgents performs script reverse-engineering to construct a hierarchical narrative memory to provide multi-level context and employs an iterative narrative planning process that refines a creative blueprint into a final compiled script. Extensive experiments demonstrate that CineAgents significantly outperforms existing methods, generating compilations with superior narrative coherence and logical coherence.

  \keywords{Video Compilation \and Multi-Agent System \and Secondary Creation \and Cinematic Production}

\end{abstract}

\section{Introduction}
\label{sec:intro}

The explosive growth of short video platforms (\eg, Tencent Video~\cite{tencent}, TikTok~\cite{tiktok}, and YouTube~\cite{youtube}) has fundamentally reshaped how people consume entertainment~\cite{lu2025multi}. 
Audiences are increasingly drawn to fast-paced short videos over traditional long-form cinematic content. 
This trend has motivated a massive community of secondary creators to adapt existing films and TV series into condensed clips, driving an urgent demand for versatile automatic video compilation approaches to streamline this process~\cite{soe2021ai, pardo2021learning, podlesnyy}.

Early instruction-driven compilation works~\cite{pardo2024generative, xiong2022transcript, truong2016quickcut, koorathota2021editing, hu2023reinforcement} predominantly focused on casual videos characterized by linear narratives. While recent works have explored the cinematic domain, they remain constrained to predefined tasks (\eg, B-roll insertion~\cite{huber2019b}, trailer generation~\cite{wang2020learning, argaw2024towards}, and highlight detection~\cite{gan2023collaborative, hive}). Consequently, the community lacks a comprehensive benchmark to evaluate cinematic compilation under versatile user instructions, limiting the development of generalized compilation systems.

\begin{figure*}[t]
    \centering
    \captionsetup{type=figure}
    \includegraphics[width=\textwidth]{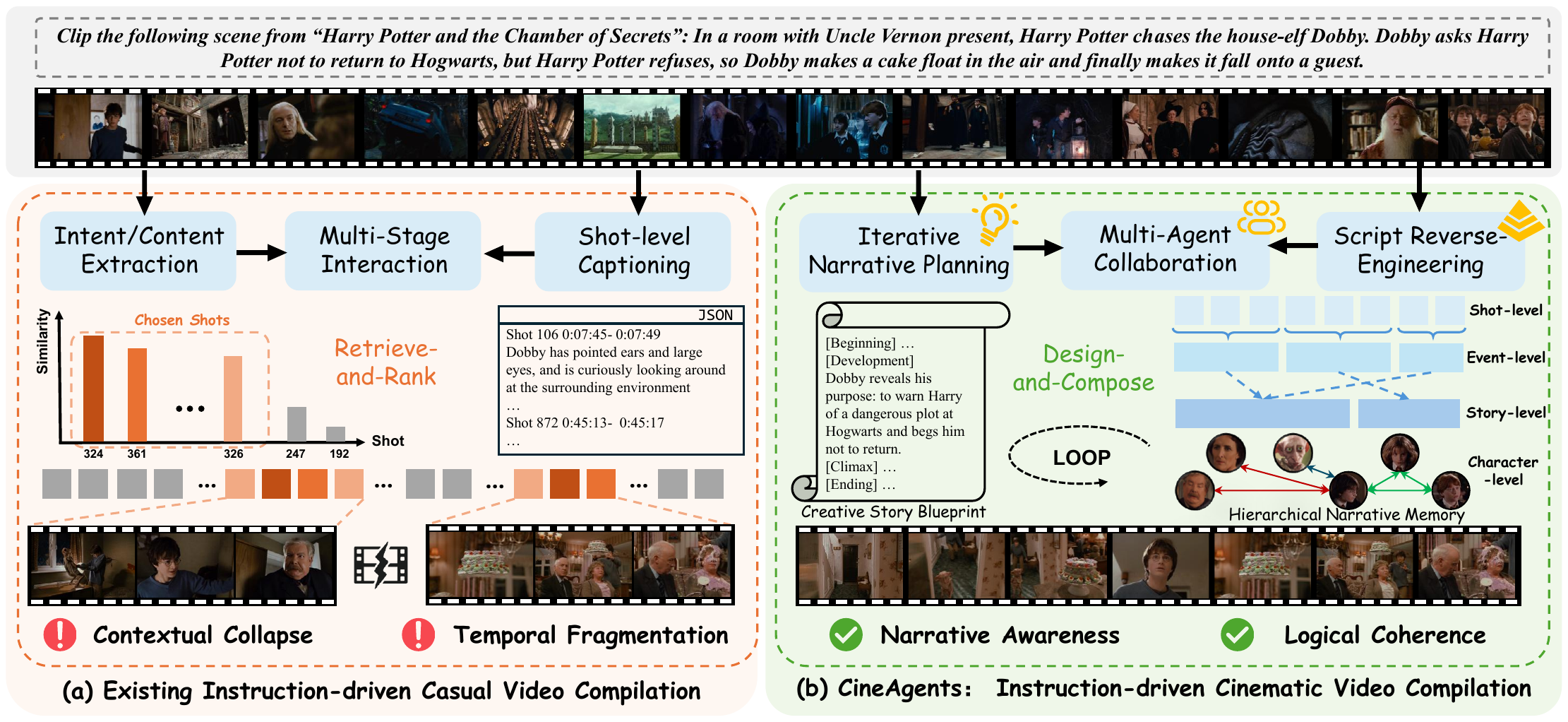}
    \captionof{figure}{Comparison of video compilation paradigms. (a) Existing video compilation methods typically rely on a ``retrieve-and-rank'' paradigm. When applied to complex cinematic videos, this isolated approach inevitably leads to contextual collapse and temporal fragmentation, leading to narratively incoherent results. 
    (b) In contrast, our proposed \textbf{CineAgents} introduces a novel ``design-and-compose'' paradigm. Through multi-agent collaboration, the system first reverse-engineers the source video into a hierarchical narrative memory, and then performs iterative narrative planning to produce a logically coherent and compelling final compilation.}
    \label{fig:teaser}
\end{figure*}

In this paper, we first propose \textbf{CineBench}, a comprehensive benchmark for the instruction-driven cinematic video compilation task. 
Unlike casual videos that typically feature linear narratives (\eg, vlogs and scenic recordings), cinematic content is built upon intricate plot structures, character arcs, and thematic undertones. 
This benchmark includes in-the-wild user instructions that range from high-level creative intents to specific editing operations, and evaluates models' abilities to perform cinematic compilation by selecting, reordering, and assembling video shots into a coherent final sequence. 
To ensure a rigorous evaluation of these complex elements, we recruited professional video editors to perform narrative re-authoring, creatively fulfilling each instruction to construct high-quality ground-truth compilations. 
As a result, CineBench contains over 500 instruction-video pairs from 70 diverse films and series, accompanied by 11 evaluation metrics organized into three core aspects.

Building upon this benchmark, we observe that existing casual video compilation methods face two key challenges:
\textit{(i) Contextual collapse}: For long cinematic videos, shot-level captions often suffer from informational redundancy and narrative inaccuracy. Consequently, models fail to grasp the broader context and retrieve shots that align with the user's instruction (\eg, \cref{fig:teaser} (a) left, the model incorrectly includes a shot of Harry and Dobby struggling over a lamp, as it over-relies on matching the characters from the prompt while ignoring the narrative context).
\textit{(ii) Temporal fragmentation}: Current methods over-rely on matching isolated content from the user's instruction, preventing them from constructing a coherent narrative flow. As a result, the generated video becomes a temporally fragmented sequence of clips lacking narrative coherence (\eg, \cref{fig:teaser} (a) right, the video abruptly cuts from Harry and Dobby's struggle to a cake falling on a guest, omitting the connecting plots).
Fundamentally, these challenges stem from their ``retrieve-and-rank'' formulation, which concatenates isolated shots based on local similarities while neglecting global narrative dependencies.

To address these limitations, we present \textbf{CineAgents}, a multi-agent system inspired by the collaboration of a professional editing studio. 
Specifically, we reformulate cinematic video compilation as a ``design-and-compose'' paradigm that orchestrates a collaborative workflow, as illustrated in \cref{fig:teaser} (b).
To overcome contextual collapse, script reverse-engineering converts the long cinematic video into a structured script. This script constructs a hierarchical narrative memory, providing narrative context by organizing the cinematic video at multiple levels (\ie, shot, event, story, and character).
To overcome temporal fragmentation, we translate the user's instruction into an initial creative story blueprint. Through an iterative narrative planning process, the system logically refines this blueprint and adjusts the shot sequence to produce the final compiled script.
Once the script meets all editing requirements, the system directly leverages external video tools to seamlessly assemble the final clips.

In summary, our contributions are listed as follows:
\begin{itemize}
    \item We construct CineBench, the first benchmark for instruction-driven cinematic video compilation, with professional editors to annotate ground truths.
    \item We present the CineAgents, a multi-agent system that reformulates instruction-driven cinematic video compilation into a design-and-compose paradigm.
    \item We introduce script reverse-engineering and hierarchical narrative memory, which analyze videos to provide narrative context across multiple levels.
    \item We propose an iterative narrative planning process that progressively refines the blueprint into compiled script, ensuring logical coherence.
\end{itemize}

\section{Related Works}
\label{sec:related}

\subsection{Casual Video Compilation}
Early research in automatic video compilation primarily relied on the similarity analysis of extracted features, leveraging audio~\cite{pavel2020rescribe, huh2023avscript}, textual~\cite{truong2016quickcut, wang2019write}, or visual~\cite{podlesnyy} cues.
Notably, CMVE~\cite{koorathota2021editing} introduced the first automatic pipeline for compilation based on a pre-defined story.
Building on this foundation, subsequent works~\cite{hu2023reinforcement, pardo2024generative, barua2025lotus} focused on enhancing editing precision by training models to ensure sequence coherence.
However, these foundational methods are inherently designed for casual videos with linear narratives, lacking the representational capacity for complex cinematic content.

\subsection{Cinematic Video Compilation}
To effectively analyze and reorganize cinematic videos, researchers have proposed a wide range of approaches, though they remain heavily confined to predefined tasks.
For instance, B-Script~\cite{huber2019b} focuses exclusively on B-roll insertion by analyzing primary footage for semantic cutaways.
Similarly, CCANet~\cite{wang2020learning} and TGT~\cite{argaw2024towards} are tailored for trailer generation by scoring segments based on attention intensity, while HIVE~\cite{hive} and CLC~\cite{gan2023collaborative} are limited to highlight detection via high-impact moment sequencing.
Since these models are constrained to specific objectives, they cannot support compilation guided by in-the-wild user instructions, preventing users from realizing their diverse creative concepts.

\subsection{Multi-Agent Cinematic Generation}
The rapid advancement of Large Language Models (LLMs)~\cite{brown2020language,gpt4,llama,llama2,towards} and Vision-Language Models (VLMs)~\cite{llava,sharegpt4v,liu2024improved,stage} has equipped AI systems with powerful reasoning and planning capabilities. 
Based on these foundation models~\cite{gpt4, gemini, claude}, recent studies have introduced LLM-driven planning to cinematic video creation. 
For instance, CineVerse~\cite{cineverse} and VideoGen-of-Thought~\cite{videogenofthought} utilize LLMs to generate structured scripts, imposing a logical narrative on the synthesized videos. 
This paradigm has further evolved into multi-agent collaboration~\cite{metagpt, zhang2024chain}, with frameworks like FilmAgent~\cite{filmagent} and MovieAgent~\cite{movieagent} proposing interactive systems for end-to-end cinematic video generation. 
However, these methods are ill-equipped for cinematic compilation, which fundamentally relies on logically reorganizing existing video shots into a coherent new narrative.

\section{Benchmark}
\begin{figure}[t]
  \centering
  \includegraphics[width=\linewidth]{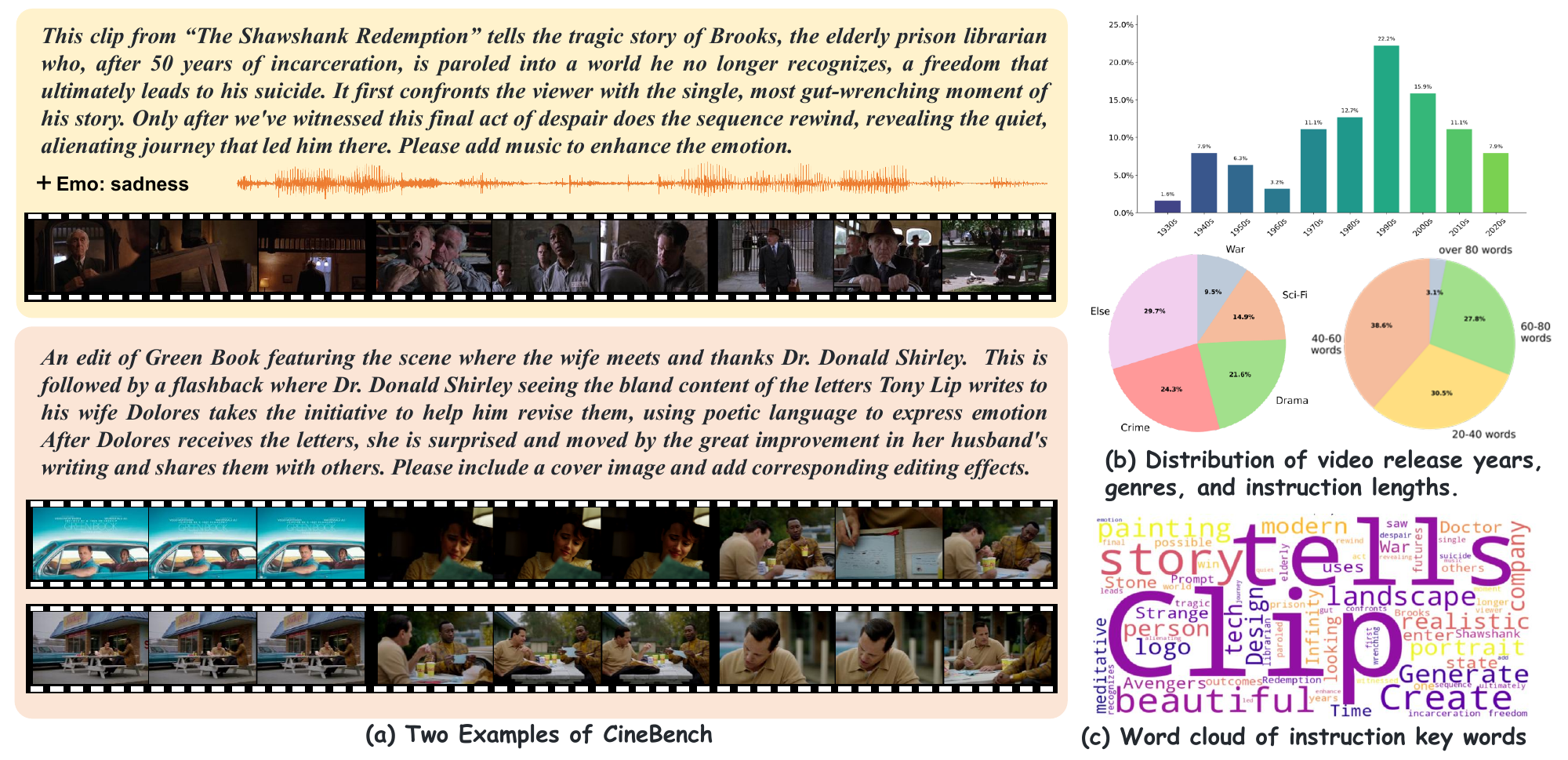}
  \caption{Overview and statistic of CineBench. (a) Examples of instructions and compiled video clips. (b) Distribution of video release years, types and instruction lengths. (c) Word cloud of instruction keyword frequencies.
  }
  \label{fig:dataset}
\vspace{-2mm}
\end{figure}

While cinematic video compilation approaches are rapidly emerging, existing benchmarks remain confined to predefined tasks (\eg, B-roll insertion~\cite{huber2019b}, trailer generation~\cite{wang2020learning, argaw2024towards}, and highlight detection~\cite{gan2023collaborative, hive}). This fundamentally limits the development of generalized compilation systems. 
To address this gap, we introduce \textbf{CineBench}\footnote{We will release the benchmark dataset upon publication.}, the first benchmark for instruction-driven cinematic video compilation, as illustrated by two representative examples in \cref{fig:dataset}~(a).

\subsection{Data Curation and Task Formulation}
To construct a comprehensive and diverse benchmark, we first curate cinematic videos from both English- and Chinese-language films and television series\footnote{These materials are used strictly for non-commercial academic research purposes.}. 
These sources span a wide range of release years (1930-2020) and genres (\eg, action, comedy, and romance) to ensure a broad representation of narrative styles.
To facilitate a rigorous evaluation of in-the-wild instructions, we deconstruct user intentions into the following fundamental components.

\noindent \textbf{Source selection.} This component defines whether the compilation draws from a single source (\eg, \textit{The Shawshank Redemption}) or integrates footage from multiple sources (\eg, combining \textit{The Shawshank Redemption} and \textit{Green Book}).

\noindent \textbf{Target content.} This component specifies the narrative focus of the compilation. This can range from a single subject, such as a character (\eg, Andy Dufresne) or an event (\eg, the prison escape), to multiple interwoven subjects, such as several interacting characters (\eg, Tony Lip and Dr. Don Shirley) or parallel plotlines (\eg, contrasting family life with solitude).

\noindent \textbf{Temporal requirements.} This component defines the temporal arrangement of clips, including preserving chronological order (\eg, a 5-minute limit within the sequence), enforcing non-linear storytelling (\eg, a highlight-first structure), and performing extractive edits (\eg, trailer generation or story summarization).

\noindent \textbf{Editing operations.} This component specifies audio-visual enhancements applied to the final compilation, such as text overlays (\eg, adding cinematic titles), background music (\eg, applying a score matching the emotional tone), cover images (\eg, using a movie poster as a title card), or transition effects (\eg, inserting fade transitions for flashback sequences).

\subsection{Ground Truth Annotation and Verification}
Guided by these four fundamental components, we recruit five professional video editors to manually craft user instructions as ground truths after watching the entirety of each source film.
To ensure the quality and consistency of these human-annotated pairs, we conduct an inter-annotator agreement analysis on a randomly selected 20\% of the data, yielding a Cohen's Kappa~\cite{kappa} of 0.61, indicating substantial agreement among the experts.
Detailed statistics regarding the distribution of cinematic videos and compilation instructions are provided in \cref{fig:dataset}~(b) and (c), encompassing release years, genres, instruction lengths, and keyword frequencies.

To further assess model robustness and evaluate hallucination in compilation frameworks, we design a challenging subset of negative instructions. 
These are systematically crafted by introducing factual inconsistencies among the source videos, characters, and plot events (\eg, requesting scenes of a character in a film in which they never appeared). 
In total, CineBench includes over 500 instruction-compilation pairs sourced from more than 70 films and series. 
Notably, 30\% samples serve as an adversarial set specifically designed to evaluate a model's ability to recognize and appropriately refuse factually impossible requests.

\subsection{Evaluation Metrics}
To enable a comprehensive evaluation of the instruction-driven cinematic video compilation, we introduce a protocol that assesses a model's performance across three key dimensions:

\noindent \textbf{Narrative grounding.} This dimension evaluates the model's comprehension of the source video's narrative. It is quantified using two metrics: inspired by Video-bench~\cite{vbench}, we compute \textit{Script-Video Consistency (SVC)} by calculating the semantic similarity between the generated script annotations and the original video at the shot level; and we employ a VLM~\cite{gemini} to rate the \textit{Shot Coherence (SC)}, assessing the logical flow between consecutive shots in the generated script.

\noindent \textbf{Retrieval precision.} This dimension assesses the accuracy and relevance of the retrieved shots. The evaluation comprises several metrics: shot-level \textit{Precision}, \textit{Recall}, and \textit{F1-score} measure selection accuracy against ground-truth annotations; the \textit{Temporal Correctness Score (TCS)} evaluates the ratio of the correctly ordered shot sequence's total duration~\cite{wagner1974string} to the generated video's total duration; and VLM-based scores calculates \textit{Narrative Logic (NL)} and \textit{Prompt Adherence (PA)}.

\noindent \textbf{Overall quality.} This dimension captures the holistic performance and reliability of the system. It is measured through three key indicators: the \textit{Execution Success Rate (ESR)} tracks the percentage of instructions processed without system errors; the \textit{Adversarial Rejection Rate (ARR)} evaluates the model's ability to correctly refuse factually impossible requests from our adversarial set; and a VLM-based \textit{Compiled Quality (CQ)} score for the final compiled video.

All VLM-based evaluations are rated on a scale of $1$ to $10$ for readability\footnote{Detailed implementation for all metrics are provided in the supplementary materials.}.

\section{Method}

\subsection{Overview}
\begin{figure}[t]
   \centering
   \includegraphics[width=\linewidth]{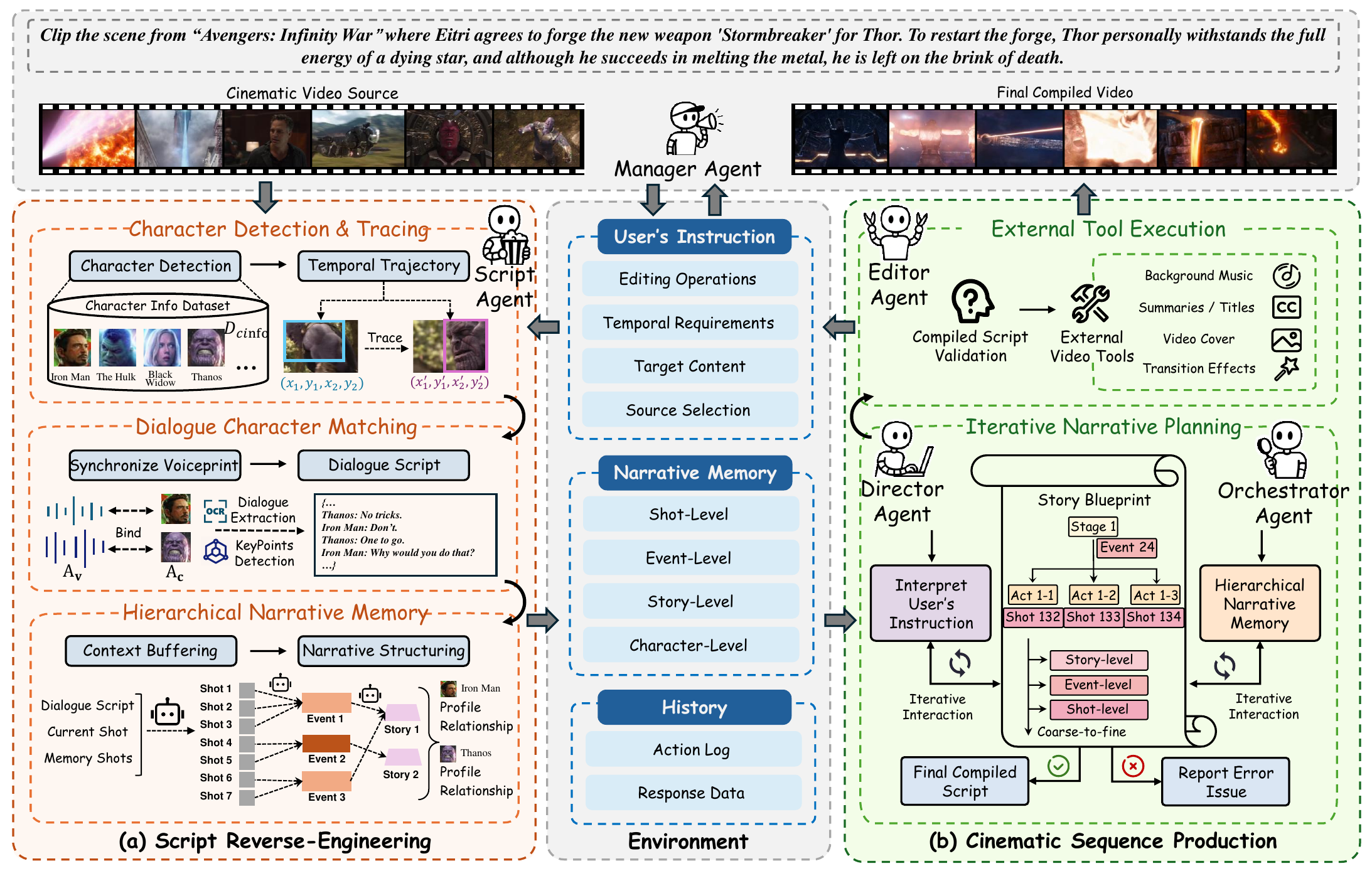}
  \caption{Overview of the proposed CineAgents. 
    Given a user instruction and a set of source videos, our system generates a final compiled video through:
    \textbf{(a) Script reverse-engineering:} The script agent integrates the multimodal information from the source videos to construct a hierarchical narrative memory, which provides a multi-level context to overcome contextual collapse (\cref{sec:script_pre}).
    \textbf{(b) Cinematic sequence production:} The director agent and orchestrator agent collaborate through iterative narrative planning to generate a compiled script to overcome temporal fragmentation. Subsequently, the editor agent applies external tools to assemble and enhance the final video (\cref{sec:production}).
    This entire collaborative pipeline is orchestrated within cinematic compilation environment, where a manager agent supervises the workflow and coordinates the specialized agents to achieve the final narrative goal (\cref{sec:environment}).
  }
  \label{fig:pipeline}
\end{figure}

We design the \textbf{CineAgents} system to generate a final compiled video $\mathcal{V}_\mathrm{out}$ from a user's text instruction $\mathcal{I}$ and a set of source videos $\mathcal{V}_\mathrm{src}$. This process is initiated by the manager agent ($\mathcal{A}_\mathrm{magr}$), which interprets the user's instruction to recruit the necessary specialist agents. As shown in \cref{fig:pipeline}, this task is accomplished through a two-stage process:

\noindent \textbf{Script reverse-engineering.}
The script agent ($\mathcal{A}_\mathrm{scr}$) first parses the source videos $\mathcal{V}_\mathrm{src}$ into a shot-level script by integrating their visual, audio, and text modalities. It then constructs a hierarchical narrative memory $\mathcal{M}$ to provide a multi-level narrative context, addressing the challenge of contextual collapse. The process can be formulated as:
\begin{equation}
    \mathcal{M} = \mathcal{A}_\mathrm{scr}(\mathcal{V}_\mathrm{src})
    \label{eq:script_memory}
\end{equation}

\noindent \textbf{Cinematic sequence production.}
This stage embodies ``design-and-compose'' paradigm to overcome temporal fragmentation. It begins with the ``design'' phase, where the director agent ($\mathcal{A}_\mathrm{dir}$) and orchestrator agent ($\mathcal{A}_\mathrm{orch}$) collaboratively perform iterative narrative planning. They transform the user's instruction $\mathcal{I}$ into a logically coherent compiled script $\mathcal{L}_\mathrm{final}$ by leveraging the narrative memory $\mathcal{M}$. This is followed by the ``compose'' phase, where the editor agent ($\mathcal{A}_\mathrm{edit}$) executes the script $\mathcal{L}_\mathrm{final}$, assembling shots from $\mathcal{V}_\mathrm{src}$ and applying external video tools to generate the final video $\mathcal{V}_\mathrm{out}$. The entire process is formulated as:
\begin{equation}
    \mathcal{V}_\mathrm{out} = \mathcal{A}_\mathrm{edit}\big((\mathcal{A}_\mathrm{orch} \leftrightarrow \mathcal{A}_\mathrm{dir})(\mathcal{I}, \mathcal{M}), \mathcal{V}_\mathrm{src}\big)
    \label{eq:production}
\end{equation}

\subsection{Script Reverse-Engineering}
\label{sec:script_pre}
The extensive length and multimodal information (text, visual, audio) of cinematic videos pose a significant challenge for existing video compilation methods, which will cause contextual collapse. 
To overcome this, we introduce Script Reverse-Engineering to transform the source videos into a hierarchical narrative memory, as shown in \cref{fig:pipeline}~(a).

\noindent\textbf{Character detection and tracing.}
We employ an ``anchor and propagate'' strategy for robust character identification.
Specifically, we establish identity anchors $\mathcal{A}_\mathrm{c}$ by creating a character dataset $\mathcal{D}_\mathrm{cinfo}$ from metadata (\eg, Wikipedia).
We perform character matching using InsightFace~\cite{insightface} for frontal views, while the SOLIDER~\cite{chen2023beyond} model handles non-frontal views where face recognition is unreliable.
We then propagate these anchor identities along temporal trajectories, which are formed by grouping shots with similar appearance features. 
This process can be formulated as:
\begin{equation}
\label{eq:char_id}
\text{ID}(T_k) = \underset{c \in \mathcal{C}}{\arg\max} \sum_{p_i \in T_k} \text{Cosine}(F(p_i), e_c),
\end{equation}
where $T_k$ is the $k$-th character trajectory composed of a set of person detections $\{p_i\}$. $F(p_i)$ is the feature embedding of a detection $p_i$ extracted by either InsightFace or SOLIDER. 
$\mathcal{C}$ is the set of all candidate characters, and $e_c \in \mathcal{A}_\mathrm{c}$ is the anchor embedding for character $c$. 
This strategy allows us to maintain consistent character identities even through occlusions and non-frontal poses.

\noindent\textbf{Dialogue character matching.}
With character identities established, we link them to their corresponding dialogues.
Since conventional ASR methods~\cite{bain2023whisperx,radford2023robust} struggle with large casts and off-screen speakers, we create voiceprint anchors $\mathcal{A}_\mathrm{v}$ from synchronized audio-visual cues. 
We extract dialogue from on-screen subtitles via OCR~\cite{ocr} and verify the on-screen speaker using lip activity detection with InsightFace~\cite{insightface} keypoints and a ResNet~\cite{resnet} classifier.
We then use WeSpeaker~\cite{wespeaker} to extract audio features from all of a character's confirmed anchors and apply K-means clustering to these features to bind a unique voiceprint $\mathcal{V}_\mathrm{c}$ to each character.
Finally, using the established speaker-dialogue mapping, we can associate each character in the cinematic source with their corresponding dialogue to generate the dialogue script.

\noindent\textbf{Hierarchical narrative memory.} 
With characters and dialogues identified, we construct the hierarchical narrative memory to distill the complex multimodal information into a multi-level structured representation.
At the shot level, we first segment the video into individual shots using AutoShot~\cite{autoshot}.
To ensure these summaries maintain narrative flow, we buffer preceding $10$ shots as memory to store key multimodal features (\eg, visual embeddings, character IDs, and dialogue) from the history context. 
An LLM~\cite{gemini} is then provided with the current shot's information along with this historical context to produce a detailed summary. 
This context-aware summarization $S_t$ for the $t$-th shot can be formulated as a conditional generation process:
\begin{equation}
\label{eq:shot_summary}
S_t = \underset{S}{\arg\max} \, P(S \mid I_t, M_t),
\end{equation}
where $S_t$ is the generated summary, $I_t$ is the current shot's information, and $M_t = \{I_{t-k}, \dots, I_{t-1}\}$ is the memory buffer.
Next, we group these summarized shots into coherent events using BaSSL~\cite{bassl}, abstract them into story, and concurrently generate character profiles. This entire hierarchical structure is encapsulated within the final narrative memory $\mathcal{M}_\text{narr}$, which is defined as:
\begin{equation}
\label{eq:narrative_memory}
\mathcal{M}_\text{narr} = \left( \mathcal{S}_\text{sum}, S_\text{event}, f_{\text{story}}(S_\text{event}), f_{\text{profile}}(S_\text{event}) \right),
\end{equation}
where $S_\text{event} = f_{\text{event}}(\mathcal{S}_\text{sum})$ and $\mathcal{S}_\text{sum} = \{S_t\}$ is the set of all shot summaries from \cref{eq:shot_summary}. The functions $f_{\text{event}}$, $f_{\text{story}}$, and $f_{\text{profile}}$ represent the processes of event grouping, story abstraction, and profile generation, respectively.

\subsection{Cinematic Sequence Production} \label{sec:production}
While the hierarchical narrative memory provides structured access to video content, naively assembling shots based on user instructions can lead to temporally fragmented and logically incoherent results. 
To overcome this challenge, we introduce the ``design-and-compose'' paradigm to plan a story blueprint and then assemble the final video, as illustrated in \cref{fig:pipeline}~(b).

\noindent \textbf{Iterative narrative planning.}
Our iterative planning module operates as a collaborative dialogue between two specialized agents. 
The director agent acts as the creative planner, translating the user's instruction into a story blueprint structured across distinct narrative stages (\eg, beginning, rising action, climax). 
Conversely, the orchestrator agent serves as the validator, tasked with grounding each proposal in concrete evidence from the hierarchical narrative memory.
Specifically, this unfolds as a top-down grounding process. 
The director agent initiates a high-level narrative intent, which the orchestrator attempts to ground at the story and character levels. 
If a consensus is reached, the blueprint is updated with the confirmed context, and the dialogue descends to the next level. 
Conversely, if supporting evidence is lacking, the orchestrator prompts the director to revise its proposal. 
This collaborative cycle of proposal, validation, and refinement can be formally expressed as an iterative update to the story blueprint $B$:
\begin{equation}
\label{eq:iterative_planning}
B_{t+1} = B_t \oplus f_{\text{orch}}(f_{\text{dir}}(U, B_t), \mathcal{M}_\text{narr}),
\end{equation}
where $B_t$ is the blueprint at iteration $t$, and $U$ is the initial user instruction. The function $f_{\text{dir}}$ represents the director agent's proposal, while $f_{\text{orch}}$ is the orchestrator agent's grounding function, which validates the proposal against the narrative memory $\mathcal{M}_\text{narr}$. The $\oplus$ operator signifies the update process: it integrates the successfully grounded evidence from $f_{\text{orch}}$ into the blueprint. If grounding fails, the result of the right-hand side is an empty set, prompting a revision by $f_{\text{dir}}$ in the next iteration. This cycle terminates when precise shots are identified to realize every request, finalizing the blueprint into a compiled script of specific shot IDs.

\noindent \textbf{External tool execution.}
Once the manager agent confirms an error-free workflow and validates the compiled script against user requirements, the editor agent initiates the final editing stage.
To fulfill the user's specified editing operations, our system provides a suite of four common external tools: adding background music, inserting text summaries or titles, generating a video cover, and applying transition effects.
By comprehensively analyzing the user's intent and the final compiled script, the editor agent selects and applies the appropriate tools, ultimately generating the final edited video sequence.

\subsection{Cinematic Compilation Environment} \label{sec:environment}
We define the cinematic compilation environment as an interactive workspace to facilitate our multi-agent system.
Within this workspace, agents operate by referring to two foundational components: the parsed user's instruction, which defines the ultimate goal of the task, and the hierarchical narrative memory, which serves as the agent's internal knowledge base of the pre-analyzed cinematic video content.
Interactions within this environment are governed by a structured message-passing protocol. Under this protocol, every agent action and its corresponding response are recorded in a shared history.
This history serves a dual purpose: it not only provides dynamic context for subsequent decisions but also acts as a version-controlled log. Crucially, this log allows the system to reload a previous compilation state for direct modification, eliminating the need to repeat foundational steps for secondary edits.
This mechanism also allows the manager agent to monitor the output of each action and make targeted interventions or responses as needed.
This design ensures that all agent actions are traceable and consistently aligned with both the user's instructions and the source materials.

\section{Experiments}

\subsection{Experiment Setup}
We evaluate our method on the proposed CineBench, a benchmark tailored for instruction-driven cinematic video compilation. 
Our CineAgents is a training-free system that leverages the powerful foundation model Gemini-2.5-pro~\cite{gemini} to orchestrate a multi-agent system. 
This system interprets a user's textual instruction and synthesizes the final compiled clips from source videos.
All experiments are conducted on a server with four NVIDIA A100 GPUs.

\begin{table}[t]
\caption{Quantitative experiment results of comparison and ablation. $\uparrow$ ($\downarrow$) means higher (lower) is better. The best performances are highlighted in \textbf{bold}.} 
\begin{center}
{
    \setlength\tabcolsep{3pt}
    \centering
    \begin{adjustbox}{width={\textwidth},totalheight={\textheight},keepaspectratio}
    \begin{tabular}{l | c c | c c c c c c | c c c}  \toprule
    \multirow{2}{*}{Method} & \multicolumn{2}{c|}{Narrative grounding} & \multicolumn{6}{c|}{Retrieval precision} & \multicolumn{3}{c}{Overall quality} \\
    & SVC $\uparrow$  & SC $\uparrow$ & Precision $\uparrow$ & Recall $\uparrow$ & F1-score $\uparrow$ & TCS $\uparrow$ & NL $\uparrow$ & PA $\uparrow$ & ESR $\uparrow$ & ARR $\uparrow$ & CQ $\uparrow$ \\ \midrule
    \multicolumn{11}{c}{Comparison with state-of-the-art methods} \cr \midrule
        
        Claude~\cite{claude} & \text{0.112} & \text{7.94} & \text{16.24} &\text{56.44} & \text{20.29} & \text{18.61\%} & \text{7.62} & \text{7.95} & \text{63.84\%} & \text{12.31\%} & \text{8.21}
        \\ 
         
        Gemini~\cite{gemini} & \text{0.131} & \text{8.57} & \text{30.88} &\text{76.21} & \text{41.59} & \text{37.41\%} & \text{8.08} & \text{8.47} & \text{74.11\%} & \text{23.85\%} & \text{8.67} 
        \\ 

        MetaGPT~\cite{metagpt} & $N/A$ & $N/A$ & \text{43.71} &\text{72.92} & \text{55.73} & \text{46.21\%} & \text{8.23} & \text{8.71} & \text{85.93\%} & \text{52.31\%} & \text{8.85}
        \\ 

        LAVE~\cite{lave} & $N/A$ & $N/A$ & \text{38.22} &\text{63.21} & \text{44.60} & \text{35.37\%} & \text{8.60} & \text{8.43} & \text{70.34\%} & \text{40.28\%} & \text{8.53} 
        \\ 
    
        VideoAgent~\cite{videoagent} & $N/A$ & $N/A$ & \text{17.86} &\text{40.05} & \text{18.84} & \text{21.35\%} & \text{7.55} & \text{7.37} & \text{47.60\%} & \text{63.31\%} & \text{7.34}
        \\ 
        
        Ours (CineAgents) & \textbf{0.154} & \textbf{9.02} & \textbf{62.43} & \textbf{76.49} & \textbf{64.13} & \textbf{52.09\%} & \textbf{8.76} & \textbf{9.06} & \textbf{92.76\%} & \textbf{87.23\%} & \textbf{9.01}
        \\ \midrule
    \multicolumn{12}{c}{Ablation study} \cr \midrule
        \textit{W/o} DCM  & \text{0.148} & \text{8.89} & \text{60.09} &\text{74.58} & \text{61.17} & \text{50.76\%} & \text{8.71} & \text{8.85} & \text{89.91\%} & \text{85.34\%} & \text{8.94}
        \cr
        \textit{W/o} HNM & $N/A$ & $N/A$ & \text{49.12} &\text{70.19} & \text{58.04} & \text{46.87\%} & \text{8.63} & \text{8.69} & \text{84.60\%} & \text{72.96\%} & \text{8.73}
        \cr
        \textit{W/o} INP & $N/A$ & $N/A$ & \text{54.70} & \text{71.67} &\text{59.52} & \text{49.27\%} & \text{8.64} & \text{8.70} & \text{85.07\%} & \text{78.81\%} & \text{8.81}
        \cr\bottomrule
    
    \end{tabular}\label{tab:comparison}
    \end{adjustbox}
}
\end{center}
\end{table}

\subsection{Comparison with State-of-the-Art Methods}
Given the absence of prior work on instruction-driven cinematic video compilation, we establish several baselines from related domains.
Our comparison experiments include: \textit{(i)} the state-of-the-art Vision Large Language Models (VLLMs) (Claude-3.7-sonnet~\cite{claude} and Gemini-2.5-pro~\cite{gemini}); \textit{(ii)} the multi-agent framework (MetaGPT~\cite{metagpt}); \textit{(iii)} the casual video compilation method (LAVE~\cite{lave}); and \textit{(iv)} the all-in-one video editing framework (VideoAgent~\cite{videoagent}).
To ensure a fair comparison, the VLLMs are prompted with a two-stage Chain-of-Thought (CoT)~\cite{chainofthought} strategy that first retrieves and then reorders shots, mirroring the retrieve-and-rank paradigm.
Moreover, both the VLLMs and MetaGPT leverage the script generated by our CineAgents, ensuring the evaluation focuses on compilation ability rather than script quality.

\begin{figure*}[t]
  \centering
  \includegraphics[width=\linewidth]{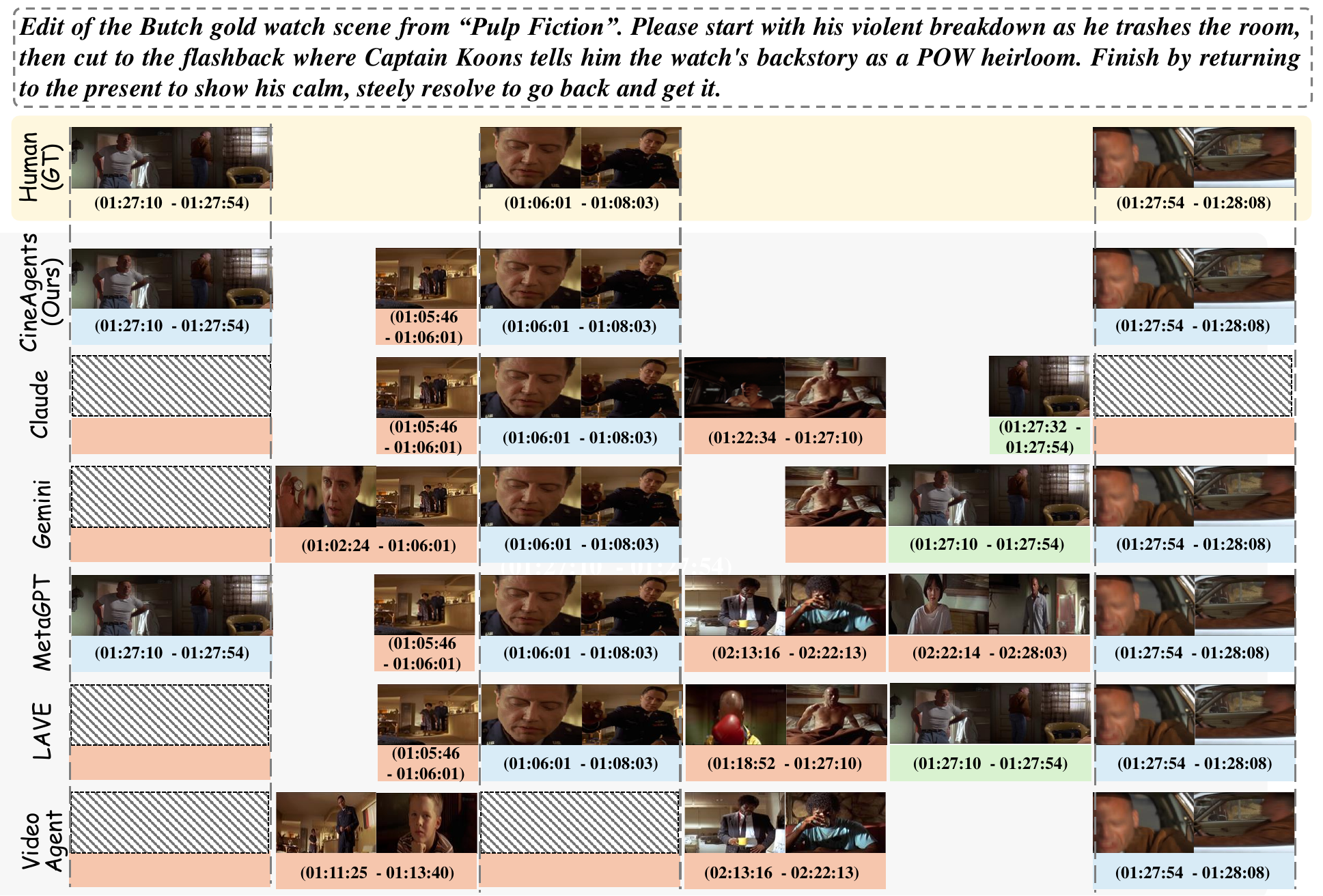}
  \caption{Qualitative comparison with state-of-the-art methods. Segments in \textbf{\textcolor{blue}{blue}} are correct in both content and temporal position, aligning with the human reference (GT); \textbf{\textcolor{ForestGreen}{green}} segments are semantically correct but temporally misplaced; \textbf{\textcolor{orange}{orange}} segments are semantically irrelevant to the instruction; and \textbf{dashed regions} represent areas where the model fails to retrieve the corresponding content.}
  \label{fig:comparison}
  \vspace{-2mm}
\end{figure*}

\noindent \textbf{Qualitative comparisons.}
We provide a visual comparison of the compiled results in \cref{fig:comparison}.
The results highlight distinct failure modes for each baseline. 
Single VLLMs like Claude~\cite{claude} and Gemini~\cite{gemini} struggle to maintain long-range narrative context, leading to shot inaccuracy. 
MetaGPT~\cite{metagpt} suffers from poor precision, retrieving numerous irrelevant shots that disrupt the narrative flow. 
LAVE~\cite{lave} is designed for casual video editing rather than cinematic video compilation, resulting in chronologically incoherent compilations. 
VideoAgent~\cite{videoagent} as an all-in-one model, lacks the specialized robustness required for the nuanced task of video compilation, making it less reliable.
In contrast, our CineAgents produces a compilation that closely mirrors the human ground truth. It successfully identifies the precise semantic events and arranges them in the exact sequence requested, generating predominantly correct segments. This visual evidence compellingly validates our system's superior capability in fine-grained narrative grounding and complex temporal reasoning.

\noindent \textbf{Quantitative comparisons.}
We present quantitative comparisons in \cref{tab:comparison}, demonstrating that CineAgents outperforms all state-of-the-art methods across all metrics. 
Specifically, CineAgent demonstrates a more accurate understanding of cinematic content (SVC), correctly interprets user instructions to retrieve corresponding clips (Precision, Recall, and F1-score), excels at arranging them in a coherent compilation sequence (TCS), and operates with high system reliability and robustness (ESR and ARR).

\begin{figure*}[t]
  \centering
  \includegraphics[width=0.8\linewidth]{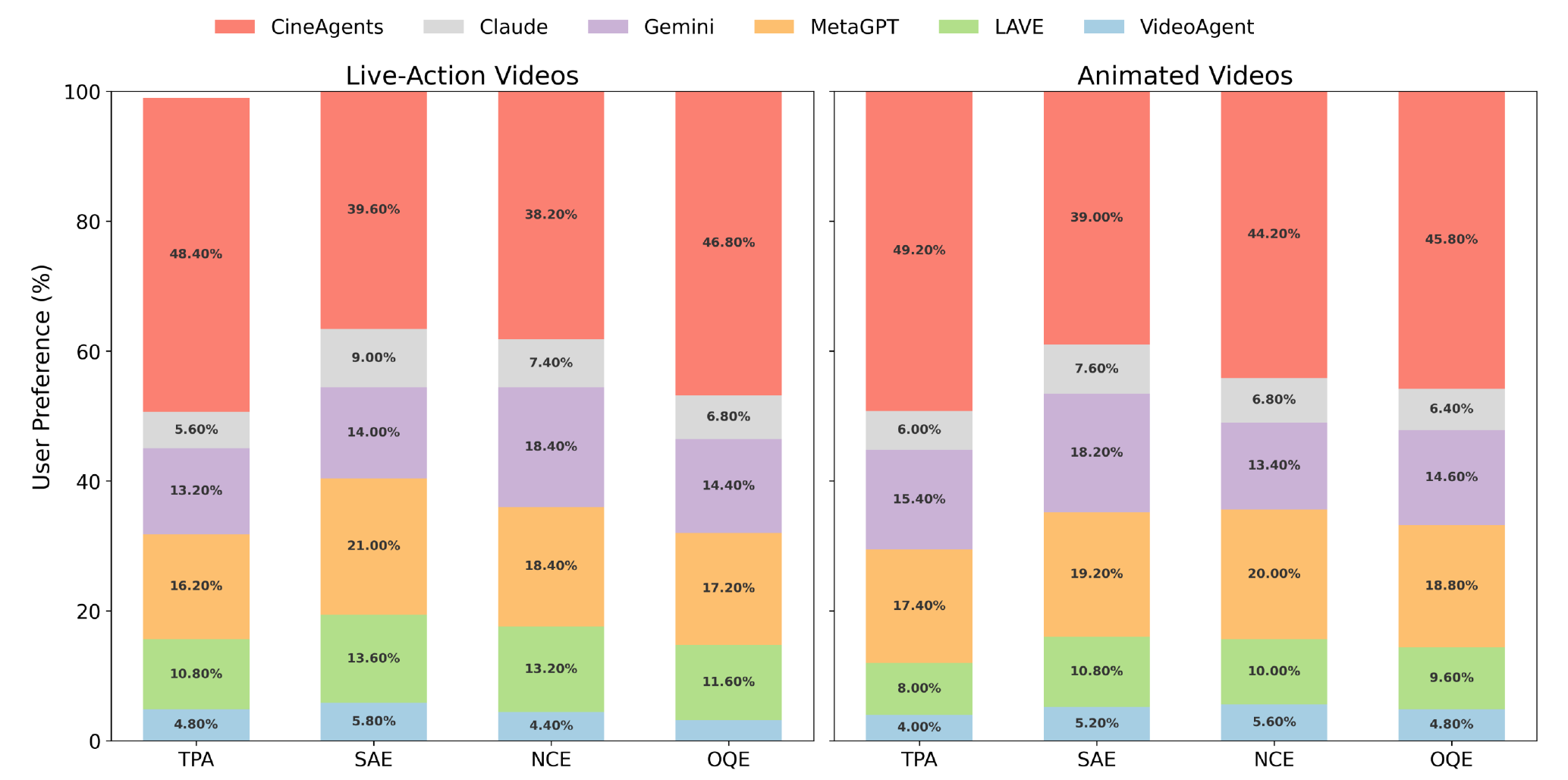}
  \caption{User study results on Live-Action and Animated videos.}
  \label{fig:user_study}
\end{figure*}

\noindent \textbf{User study.}
Given the subjective nature of cinematic video compilation, quantitative metrics and LLM-based evaluations are insufficient for a comprehensive assessment. 
Therefore, we conduct four user studies to evaluate human preference for our results:
\textit{(i)} \textbf{Text-Prompt Alignment (TPA):} Participants are asked to select the video that most faithfully represents the content of the user's instruction.
\textit{(ii)} \textbf{Sequence Alignment Evaluation (SAE):} Participants are tasked with identifying the compilation whose shot sequence most accurately follows the order specified in the instruction.
\textit{(iii)} \textbf{Narrative Coherence Evaluation (NCE):} This study assesses internal consistency, asking participants to choose the video that presents the most logical and coherent story.
\textit{(iv)} \textbf{Overall Quality Evaluation (OQE):} Participants select the best video, considering all factors including content relevance, narrative flow, and overall viewing experience.
For each study, we randomly selected 50 samples from CineBench and recruited 25 volunteers to provide independent evaluations. 
As shown in \cref{fig:user_study}, our model achieves the highest preference scores across all four evaluations.

\subsection{Ablation study}
We discard various modules and create three variants to study the impact of our proposed modules. The scores of the ablation study are shown in \cref{tab:comparison}.

\noindent \textbf{W/o DCM (Dialogue Character Matching).}
We discard the dialogue character matching module, which prevents the model from associating dialogue with specific characters. As a result, the model lacks the character-dialogue cues to accurately annotate the script (lower SVC score).

\noindent \textbf{W/o HNM (Hierarchical Narrative Memory).}
We discard the hierarchical narrative memory module, forcing the model to rely solely on shot-level information. This prevents the system from building high-level understanding of the narrative, causing inaccurate and irrelevant retrieval results (lower F1-score).

\noindent \textbf{W/o INP (Iterative Narrative Planning).}
We replace the iterative narrative planing with a retrieve-and-rank method. By prioritizing superficial keyword similarity, this variant loses the ability to distinguish between feasible and unfeasible requests. As a result, when challenged with an adversarial set, it fails to refuse instructions (lower ARR score).

\section{Discussion}

\subsection{Different Task Clarification}
We clarify the differences between video compilation and related tasks. These tasks operate at different levels:

\begin{itemize}
    \item Video generation and editing are concerned with pixel-level manipulation. They focus on synthesizing visual content or altering the pixels of frames.
    \item Video understanding involves the passive analysis of video content. Its goal is to interpret semantics for tasks like question answering, without any capability to modify or rearrange the video structure.
    \item Video compilation operates at the structural level which selects and sequences existing video clips to construct a new narrative. It is worth noting that some works (\eg, EditDuet~\cite{editduet} and HIVE~\cite{hive}) are not open-source.
\end{itemize}

\subsection{Generalization Across Genres}
To quantify this generalization, we conduct four user studies on 10 animated films and 50 instructions. 
As detailed in \cref{fig:user_study}, participants evaluate the results using the same four criteria (TPA, SAE, NCE, and OQE), with the results confirming our model's superior versatility across genres.

\begin{figure*}[t]
  \centering
  \includegraphics[width=\linewidth]{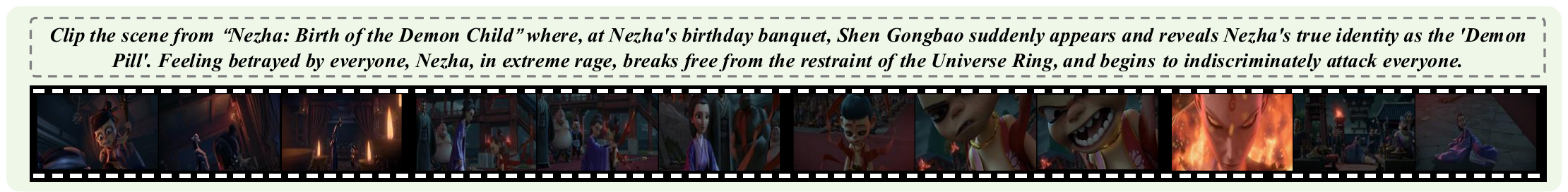}
  \caption{Visual example of CineAgents compilation performance on animated content. This highlights the model's generalization capability beyond live-action video.}
  \label{fig:discusion}
  \vspace{-2mm}
\end{figure*}

\subsection{Efficiency and Cost Analysis}
Transforming a video into a hierarchical narrative memory for a film or TV series takes an average of $1$ hour and costs around \$$14$ in API fees via the Gemini-2.5-pro~\cite{gemini}. 
This comprehensive memory is fully reusable for all subsequent instructions. 
Consequently, Executing a single instruction takes an average of $6$ minutes and costs approximately \$$0.62$ in API fees.
This highlights the system's practicality for iterative, real-world creative workflows.

\section{Conclusion}
In this paper, we present CineBench, the first comprehensive benchmark for instruction-driven cinematic video compilation, featuring high-quality ground truths crafted by professional video editors. 
To tackle this challenging task, we introduce CineAgents, a multi-agent system that reformulates the traditional retrieve-and-rank approach into a design-and-compose paradigm. 
Specifically, our system employs script reverse-engineering to construct a hierarchical narrative memory, effectively overcoming contextual collapse. Furthermore, it utilizes an iterative narrative planning process to ensure the logical coherence, resolving temporal fragmentation. 
Extensive experiments demonstrate that CineAgents significantly outperforms existing methods in narrative grounding, temporal correctness, and alignment with complex user instructions. 
We believe CineBench and CineAgents represent a significant step toward generalized compilation systems, empowering creators to efficiently compile cinematic content.

\noindent \textbf{Limitations.}
While CineAgents demonstrates superior performance in cinematic video compilation, its current implementation relies on off-the-shelf computer vision models (\eg, InsightFace and SOLIDER) for character detection and temporal tracing. In complex cinematic scenes (\eg, extreme lighting, motion blur, and view occlusions), these models may fail, and such tracking inaccuracies can cascade into errors within the hierarchical narrative memory. 
We leave the integration of more advanced perceptual modules as future work to further improve the compilation accuracy of CineAgents.

%
%
\bibliographystyle{splncs04}
\bibliography{main}
\end{document}